# REDUCTIVE PROPERTY OF NEW FUZZY REASONING METHOD BASED ON DISTANCE MEASURE

SON-IL KWAK*, GUM-JU KIM, MICHIO SUGENO, GWANG-CHOL LI, MYONG-SUK SON, HYOK-CHOL KIM, UN-HA KIM

**ABSTRACT** Firstly in this paper we propose a new criterion function for evaluation of the reductive property about the fuzzy reasoning result for fuzzy modus ponens and fuzzy modus tollens. Secondly unlike fuzzy reasoning methods based on the similarity measure, we propose a new fuzzy reasoning method based on distance measure. Thirdly the reductive property for 5 fuzzy reasoning methods are checked with respect to fuzzy modus ponens and fuzzy modus tollens. Through the experiment, we show that proposed method is better than the previous methods in accordance with human thinking.

## 1. Introduction

Fuzzy modus ponens (FMP) and fuzzy modus tollens (FMT) are two fundamental pattern of general fuzzy reasoning [31]. Zadeh in [31] proposed the Compositional Rule of Inference (CRI) for FMP and FMT. Wang in [25] presented the Triple Implication Principle (TIP) with total inference rules of fuzzy reasoning. Since the inception of the triple I method [25], many papers have researched the fuzzy inference method. The paper [29] showed that results about the triple implication method for fuzzy reasoning obtained in [5] are correct and the Example 2.1 in [5] is incorrect. In [33] the quintuple implication principle (QIP) for solving FMP and FMT was proposed unlike Zadeh's CRI [31] and Wang's TIP [17]. Reductive Property is one of the essential and important properties in the applications of the fuzzy inference mechanism [17, 29]. Since the inception of Zadeh's pioneering paper [31], the reductive property of the fuzzy reasoning method in the fuzzy inference systems (FIS) has been very important topic in the fuzzy theory and its applications [17, 27, 7, 20]. In [22] fuzzy similarity inference (FSI) is proposed, first, concept of the fuzzy similarity measure is axiomatically defined. It is a generalization of the similarity measure. Several formulas are presented to calculate the fuzzy similarity of two fuzzy sets. Second, in [22], a fuzzy reasoning method is presented, which is called fuzzy similarity inference, and computational formulas for both FMP and FMT are drowned. And then computational formulas for $\alpha$-FSI FMP and $\alpha$-FSI FMT are proposed based on [1, 18, 24]. Finally, reductive, i.e., reversibility properties of the proposed FSI are discussed. In [26] is pointed out that many papers have been done for computing and analyzing the fuzzy inference conclusion $B^*$ which are valuable for solving problems in fuzzy control and are meaningful with respect to theoretic aspect. On the other hand, according to [2, 6, 12-15], it mathematically seems that they are all accompanied with a common shortcoming, that is, information loss. In [30] reverse triple I method of fuzzy reasoning was proposed. In [11] based on Schweizer–Sklar operators new triple I algorithms was presented. In [30] based on Lukasiewicz implication operator the reverse triple I method was proposed. In [32] the triple I method of fuzzy reasoning based on intuitionistic fuzzy set is presented. However in [8] shortcoming of the triple I method is pointed out, which cannot be applied in fuzzy control. There are a lot of the fuzzy inference method based on similarity measures. In [20, 21], the shortcomings of CRI method are mentioned, so a similarity-based fuzzy reasoning method called approximate analogical reasoning schema (AARS) is proposed. In [3, 4] in order to solve the problems in medical diagnosis, two similarity-based fuzzy reasoning methods called MF based on matching function and FT based on function T are proposed. In [28] three fuzzy reasoning methods based on similarity, i.e., IC based on inclusion and cardinality, DS based on degree of subsethood, and EC based on equality and cardinality are proposed. In [23] a fuzzy similarity inference scheme is proposed, which is to extend from the general similarity measure to fuzzy similarity measure. The principle of fuzzy reasoning methods based on similarity is to obtain the fuzzy set $B^*$ of the inference result by modifying fuzzy set $B$ of the consequent with a modification function based on the similarity between $A$ and $A^*$. Compared with the CRI method, the fuzzy reasoning methods based on similarity do not require the calculation of fuzzy relation. However the results obtained by these methods, strongly depend on the similarity measure and the modification function. In [16] author pointed out that the fuzzy reasoning methods based on the fuzzy relation $R_m$, $R_a$, $R_c$, and $R_p$ do not satisfy the reductive property, but they can be applied to the practical problem, for example fuzzy control. The authors mentioned in [13-16] that the reasoning methods based on the fuzzy relation $R_{ss}$, $R_{sg}$, $R_s$, $R_{gg}$, $R_{gs}$, and $R_g$ do satisfy the reductive property, but they cannot be applied to the practical problem, for example fuzzy control. That is, as mentioned in [16], this is contradict. The experimental fuzzy control results are presented in [16].

To overcome this contradiction between the reductive property and fuzzy control, fuzzy reasoning method based on a new principle must be developed, so far, a lot of fuzzy reasoning methods were proposed, studied, checked, and applied in many branches.

In this paper we propose a new criterion function for checking of the reductive property about the fuzzy reasoning result for fuzzy modus ponens and fuzzy modus tollens. And then, unlike fuzzy reasoning methods based on the similarity measure, we propose a new fuzzy reasoning method based on distance measure (DM) presented in [20]. And then we compare with the reductive properties for 5 fuzzy reasoning methods are compared with respect to fuzzy modus ponens and fuzzy modus tollens, which are CRI, TIP, QIP, AARS, and an our new DM method. We show that DM method proposed in this paper is better in accordance with human thinking than CRI, TIP, AARS and QIP.

The rest of this paper is organized as follows. In section 2, we discuss backgrounds of CRI, TIP, QIP, and AARS for FMP and FMT. And we present about the fuzzy reasoning methods based on fuzzy relation. In section 3, a new criterion function and fuzzy reasoning methods for FMP and FMT are presented, respectively. In section 4, the reductive property of CRI, TIP, and QIP for FMP and FMT is checked by using Łukasiewicz's implication and the corresponding t-norm, Gödel's implication and the corresponding t-norm, $R_0$ implication and the corresponding t-norm, and Gougen's implication and the corresponding t-norm, the reductive property of AARS for FMP and FMT is checked by using more or less form and reduction form, and then the reductive property of our method is checked by using new 2 forms. And then five fuzzy reasoning methods, i.e., Zadeh's CRI method, G. J. Wang's TIP method, Baokui Zhou et al.'s the QIP method, I. B. Turksen et al.'s AARS method, and DMM proposed in this paper are compared with respect to the reductive property.

## 2. Backgrounds

Generally known fuzzy reasoning are FMP and FMT in the fuzzy system with 1 input 1 output 1 rule. General form of the fuzzy modus ponens in [5] is as follows.

$$\begin{aligned} &\text{Rule};\quad &&\textit{if }x\textit{ is }A\textit{ then }y\textit{ is }B\\ &\text{Premise}:\quad &&x\textit{ is }A^*\\ &\text{Conclusion}:\quad &&y\textit{ is }B^* \end{aligned} \qquad(1)$$

General form of FMT in the paper [5] is as follows.

$$\begin{aligned} &\text{Rule};\quad &&\textit{if }x\textit{ is }A\textit{ then }y\textit{ is }B\\ &\text{Premise}:\quad &&y\textit{ is }B^*\\ &\text{Conclusion}:\quad &&x\textit{ is }A^* \end{aligned} \qquad(2)$$

, where $A^* \in F(X)$, $A \in F(X)$ are fuzzy sets defined in the universe of discourse $X$, $B^* \in F(Y)$, $B \in F(Y)$ are fuzzy sets defined in the universe of discourse $Y$. In the fuzzy system with m input 1output n rules, we rewrite the definition for reductive property of fuzzy inference method in [5]. According to [10], the formula (2) can be written as follows, because FMT is opposite with FMP.

$$\begin{aligned} &\text{Rule};\quad &&\textit{if }y\textit{ is }\overline{B}\textit{ then }x\textit{ is }\overline{A}\\ &\text{Premise}:\quad &&y\textit{ is }B^*\\ &\text{Conclusion}:\quad &&x\textit{ is }A^* \end{aligned} \qquad(3)$$

For formula (1), (2), and (3), according to Zadeh's viewpoint, Rule is represented by some fuzzy relation. For example, when $\rightarrow_z$ is Zadeh's implication, the fuzzy relation of the Rule is presented as follows.

$$\begin{aligned} &R(x, y) = A(x) \rightarrow_z B(y)\\ &a \rightarrow_z b = (1-a) \vee (a \wedge b) \end{aligned} \qquad(4)$$

In [33], authors listed four most important implication operators and the corresponding t-norms. As mentioned in [33], Łukasiewicz's implication $a \rightarrow_L b$ and the corresponding t-norm $a \otimes_L b$, Gödel's implication $a \rightarrow_G b$ and the corresponding t-norm $a \otimes_G b$, $R_0$ implication $a \rightarrow_{R_0} b$ and the corresponding t-norm $a \otimes_{R_0} b$, and Gougen's implication $a \rightarrow_{G_0} b$ and the corresponding t-norm $a \otimes_{G_0} b$ are as follows, respectively.

$$a \rightarrow_L b = 1 \wedge (1-a+b), \qquad a \otimes_L b = 0 \vee (a+b-1) \qquad(5)$$

$$a \to_G b = \begin{cases} 1, & \text{if } a \le b \\ b, & \text{if } a > b \end{cases} \qquad a \otimes_G b = a \wedge b \qquad (6)$$

$$a \to_{R_0} b = \begin{cases} 1, & \text{if } a \le b \\ a' \vee b, & \text{if } a > b \end{cases} \qquad a \otimes_{R_0} b = \begin{cases} 0, & \text{if } a+b \le 1 \\ a \wedge b, & \text{if } a+b > 1 \end{cases} \qquad a' = 1-a \qquad (7)$$

$$a \to_{G_0} b = \begin{cases} 1, & \text{if } a \le b \\ \dfrac{b}{a}, & \text{if } a > b \end{cases} \qquad a \otimes_{G_0} b = ab \qquad (8)$$

## 3. New Reductive Property Criterion Function and Fuzzy Reasoning Method

In this section, we newly propose a criterion function for evaluation of the reductive property about the fuzzy reasoning result. And then, unlike fuzzy reasoning method based on similarity measure [28], and we propose two new fuzzy reasoning methods based on distance measure (DM) presented in [21].

### 3. 1. Motivation and Importance of New Fuzzy Reasoning Method
1) Motivation

In a lot of papers fuzzy reasoning methods based on similarity measure are proposed. Their basic idea is that the reductive property should consider the similarity measure of the consequent $B(y)$ and the fuzzy reasoning conclusion $B^*(y)$ if the antecedent $A(x)$ is similar to the given premise $A^*(x)$ for FMP. This idea is right. By the way "the antecedent $A(x)$ is similar to the given premise $A^*(x)$" is approximately equal to "the antecedent $A(x)$ is closer to the given premise $A^*(x)$". Here "similar" is correspondent to similarity measure, "closer" to distance measure. Similarity measure and distance measure have inverse proportional relation. That is, if $A(x)$ is completely equal to $A^*(x)$ then the similarity measure is 1 and distance measure is 0. Based on distance measure [11] mentioned above, in this paper we attempt to propose a new fuzzy reasoning method based on Turksen and Zhong's Euclidian distance measure(DM), so called DMM, which consists of both DM for FMP and DM for FMT, for short, FMP-DM, and FMT-DM, respectively. And the similarity measure has closed interval [0, 1] and distance measure [0, $m$], where $m$ is a finite number, $m > 0$. Using this fact fuzzy reasoning based on distance measure is possible. This is a motivation of this paper.

2) Possibility and Importance

The fuzzy reasoning methods based on similarity do not require the calculation of fuzzy relation or implication. However the results obtained by the similarity methods depend strongly on the similarity measure and the modification function. Therefore fuzzy reasoning method that does not depend on the similarity measure and can satisfy the reductive property must be researched. The fuzzy reasoning methods based on similarity do use nonlinear, i.e., max, min, product, and division operator. Thus fuzzy reasoning methods based on similarity measure have a lot of information loss [26]. But fuzzy reasoning methods based on distance measure can use linear operator for example summation and subtraction, thus information loss can be reduced. This is a possibility and an importance of this paper.

### 3.2. Reductive Property Criterion Function

The reductive property is one of the essential properties in the applications of the fuzzy inference mechanism [5, 16]. Reductive property for FMP and FMT is shown in Table 1.

According to [16, 28], four cases of the Premise for FMP in Class 1 are as follows; Case 1: $A^*$ is $A$, Case 2: $A^*$ is very $A$, Case 3: $A^*$ is more or less $A$, and Case 4: $A^*$ is not $A$. Since FMT is opposite with FMP, according to [16, 28], four cases of the given premise for FMT in Class 1 are as follows; Case 6: $B^*$ is not $B$, Case 7: $B^*$ is not very $B$, Case 8: $B^*$ is not more or less $B$, and Case 9: $B^*$ is $B$. And four cases of the Premise for FMP in Class 2 are as follows; Case 1: $A^*$ is $A$, Case 2: $A^*$ is very $A$, Case 3: $A^*$ is more or less $A$, and Case 5: $A^* =$ slightly tilted of $A$, and four cases of the given premise for FMT in Class 2 are as follows; Case 6: $B^*$ is not $B$, Case 7: $B^*$ is not very $B$, Case 8: $B^*$ is not more or less $B$, and Case 10: $B^* =$ slightly tilted of $B$. What conclusion B* for FMP and A* for FMT can be obtain? For this problem, Table 1 shows Reductive Property of FMP and FMT based on [5, 16]. In Table 1, Case 4 is a criterion based on the paper [16], Case 10 a criterion based on the paper [5], for FMP and FMT, respectively. In other words, Case 1, Case 2, Case 3, and Case 4 are criterion functions based on the paper [16], Case 6, Case 7, Case 8, and Case 10 are criterion functions based on the paper [5]. In the paper [5], authors mentioned that their proposed method is based on the assumption that the premise $A^*$ is slightly different from the antecedent of fuzzy rule $A$ and thus the conclusion $B^*$ is slightly different from the consequent $B$ of fuzzy rule, therefore, they do not expect a reasonable conclusion if the premise $A^*$ is different from the antecedent A too much. Unlike the classical reasoning, if the given premise $A^*$ is not exactly equal to the antecedent $A$, we can still obtain fuzzy reasoning result $B^*$. However we know that if the given premise $A^*$ and the antecedent $A$ are totally different, then the fuzzy reasoning result $B^*$ might be unreasonable or uninformative. Then in practical applications, a group of fuzzy rules called rule base is used to avoid the incorrect fuzzy reasoning result caused by the deviation between the given premise $A^*$ and the antecedent $A$. As obviously mentioned in the paper [9], if the given premise $A^*$ is slightly different from the antecedent $A$ then the fuzzy reasoning conclusion $B^*$ is slightly different from the consequent $B$. According to combination of the paper [5] and [16], for example the antecedent fuzzy set $A = [small]$ and consequent fuzzy set $B = [large]$, we can obtain the following Table 1. Class 1 and Class 2 are as follows in Table 1. That is, Class 1; Case 1, Case 2, Case 3, and Case 4 for FMP, and Case 6, Case 7, Case 8, and Case 9 for FMT. Class 2; Case 1, Case 2, Case 3, and Case 5 for FMP, and Case 6, Case 7, Case 8, and Case 10 for FMT. Since FMT is opposite FMP, Case 1 corresponds to Case 6, Case 2 to Case 7, Case 3 to Case 8, Case 4 to Case 9, and Case 5 to Case 10, respectively. In [5], Conclusion "or $B^* = B = [large]$" in Case 2 and Case 3 for FMP is absented, also, Conclusion "or $A^* = $ not $A = 1-[small]$" in Case 7 and Case 8 for FMT is absented. And in [33] according to Table 1, Case 2, Case 3, Case 4 and Case 5 for FMP are absented, also, Case 6, Case 7, Case 8 and Case 10 for FMT are absented, with respect to [5, 16].

**Table 1.** Reductive Property for FMP and FMT obtained by [5, 16]

| FMP | *if x is A then y is B* (Rule) | |
|---|---|---|
| | *x is $A^*$* (Premise) | *y is $B^*$* (Conclusion) |
| Case 1 | $A^* = A = [small]$ | $B^* = B = [large]$ |
| Case 2 | $A^* = $ *very* $A = [small]^2$ | $B^* = $ *very* $B = [large]^2$ or $B^* = B = [large]$ |
| Case 3 | $A^* = $ *more or less* $A = [small]^{1/2}$ | $B^* = $ *more or less* $B = [large]^{1/2}$ or $B^* = B = [large]$ |
| Case 4 | $A^* = $ *not* $A = 1-[small]$ | $B^* = $ *not* $B = 1-[large]$ |
| Case 5 | $A^* = $ *slightly tilted of* $A$ | $B^* = $ *slightly tilted of* $B$ |
| FMT | *if y is $\overline{B}$ then x is $\overline{A}$* | |
| | *y is $B^*$* (Premise) | *x is $A^*$* (Conclusion) |
| Case 6 | $B^* = $ *not* $B = 1-[large]$ | $A^* = $ *not* $A = 1-[small]$ |
| Case 7 | $B^* = $ *not very* $B = 1-[large]^2$ | $A^* = $ *not very* $A = 1-[small]^2$ or $A^* = $ *not* $A$ |
| Case 8 | $B^* = $ *not more or less* $B = 1-[large]^{1/2}$ | $A^* = $ *not more or less* $A$ or $A^* = $ *not* $A$ |
| Case 9 | $B^* = B = [large]$ | $A^* = A = [small]$ |
| Case 10 | $B^* = $ *slightly tilted of* $B$ | $A^* = $ *slightly tilted of* $A$ |

The checking function for reductive property can be defined as the difference between the consequent of fuzzy rule and conclusion of the fuzzy reasoning. For this, several concepts are defined as follows.

**Definition 3.1.** *Let fuzzy sets $A \in F(X)$, $A_l^* \in F(X)$, $B \in F(Y)$ and $B_l^* \in F(Y)$, ($l=1, 2, \cdots, s$, $k=1, 2, ..., r$), for FMP be their antecedent vectors $A = [a_1, a_2, ..., a_k, ..., a_r]$, the given premise vector $A_l^* = [a_{1l}^*, a_{2l}^*, ..., a_{kl}^*, ..., a_{rl}^*]$, and the consequent vector $B = [b_1, b_2, ..., b_k, ..., b_r]$. And then let the fuzzy reasoning conclusion be $B_l^* = [b_{1l}^*, b_{2l}^*, ..., b_{kl}^*, ..., b_{rl}^*]$. Then the error $E(B_l^*, B)$ between the conclusion $B_l^*$ and consequent $B$, and the error $e(A_l^*, A)$ between the given premise $A_l^*$ and the antecedent $A$ are defined as follows, respectively.*

$$E(B_l^*, B) = [b_{1l}^*, b_{2l}^*, ..., b_{kl}^*, ..., b_{rl}^*] - [b_1, b_2, ..., b_k, ..., b_r]$$
$$e(A_l^*, A) = [a_{1l}^*, a_{2l}^*, ..., a_{kl}^*, ..., a_{rl}^*] - [a_1, a_2, ..., a_k, ..., a_r] \quad (9)$$

**Definition 3.2.** *This Definition 3.2 is to generalize of the criterion for FMP shown in Table 1 according to [2, 4, 20-22]. The lth reductive property criterion function $RPCF_{FMP-FR-I}^l$ for the Case l (l=1, 2, 3, 4, and 5, from Table 1) in FMP can be illustratively defined as follows.*

$$RPCF_{FMP-FR-I}^{l} = \begin{cases} (1-\sum_{k=1}^{r}|b_{kl}^{*}-b_{k}|/r)\times 100, & \text{for Case 1} \\ (1-\sum_{k=1}^{r}|b_{kl}^{*}-b_{k}|/r)\times 100, \text{ or } (1-\sum_{k=1}^{r}|b_{kl}^{*}-b_{k}^{2}|/r)\times 100, & \text{for Case 2} \\ (1-\sum_{k=1}^{r}|b_{kl}^{*}-b_{k}|/r)\times 100, \text{ or } (1-\sum_{k=1}^{r}|b_{kl}^{*}-b_{k}^{\frac{1}{2}}|/r)\times 100, & \text{for Case 3} \\ (1-\sum_{k=1}^{r}|b_{kl}^{*}-(1-b_{k})|/r)\times 100, & \text{for Case 4} \\ (1-\sum_{k=1}^{r}|b_{kl}^{*}-slightly\ tilted\ of\ b_{k}|/r)\times 100, & \text{for Case 5} \end{cases}, (\%) \quad (10)$$

In Case 5 the given premise is $A^{*} = slightly\ tilted\ of\ A$, and Conclusion $B^{*} = slightly\ tilted\ of\ B$.

**Definition 3.3.** *The reductive property criterion function* $RPCF_{FMP-FR}$ *for FMP of a fuzzy reasoning method (or algorithm) is defined as follows.*

$$RPCF_{FMP-FR} = \frac{1}{S}\sum_{l=1}^{S} RPCF_{FMP-FR}^{l}, (\%) \quad (11)$$

According to Definition 3.3 and Table 1, Class 1 containsw Case 1, 2, 3, and 4 for FMP, and Case 6, 7, 8, and 9 for FMT, and then Class 2 contains Case 1, 2, 3, and 5 for FMP, Case 6, 7, 8, and 10, for FMT, therefore $s$ is 4.

**Definition 3.4.** *Since FMT is opposite to FMP, let us consider the formula (3) instead of formula (2) for FMT. Now let fuzzy sets* $\overline{B} \in F(Y)$, $B_{l}^{*} \in F(Y)$, *and* $\overline{A} \in F(X)$ *be antecedent vectors* $\overline{B}=[1-b_{1}, 1-b_{2}, ..., 1-b_{k}, ..., 1-b_{r}]$, *the given premise vector* $B_{l}^{*} = [b_{1l}^{*}, b_{2l}^{*}, ..., b_{kl}^{*}, ..., b_{rl}^{*}]$, *and the consequent vector* $\overline{A} = [1-a_{1}, 1-a_{2}, ..., 1-a_{k}, ..., 1-a_{r}]$ *of fuzzy rule. And the conclusion* $A_{l}^{*} \in F(X)$ *be* $A_{l}^{*} = [a_{1l}^{*}, a_{2l}^{*}, ..., a_{kl}^{*}, ..., a_{rl}^{*}]$, $(l=1, 2, \cdots, s, k=1, 2, ..., r)$. *Then for FMT the error* $E(A_{l}^{*}, \overline{A})$ *between the fuzzy reasoning conclusion* $A_{l}^{*}$ *and consequent* $\overline{A}$ *of fuzzy rule, and the error* $e(B_{l}^{*}, \overline{B})$ *between the given premise* $B_{l}^{*}$ *and their antecedent* $\overline{B}$ *are defined as follows, respectively.*

$$E(A_{l}^{*}, \overline{A}) = [a_{1l}^{*}, a_{2l}^{*}, ..., a_{kl}^{*}, ..., a_{rl}^{*}] - [1-a_{1}, 1-a_{2}, ..., 1-a_{k}, ..., 1-a_{r}]$$
$$e(B_{l}^{*}, \overline{B}) = [b_{1l}^{*}, b_{2l}^{*}, ..., b_{kl}^{*}, ..., b_{rl}^{*}] - [1-b_{1}, 1-b_{2}, ..., 1-b_{k}, ..., 1-b_{r}] \quad (12)$$

**Definition 3.5.** *The lth reductive property criterion function* $RPCF_{FMT-FR-I}^{l}$ *for the Case l (l=6, 7, 8, 9, and 10, from Table 1) in FMT can be illustratively defined as follows.*

$$RPCF_{FMT-FR-I}^{l} = \begin{cases} (1-\sum_{k=1}^{r}|a_{kl}^{*}-(1-a_{k})|/r)\times 100, & \text{for Case 6} \\ (1-\sum_{k=1}^{r}|a_{kl}^{*}-(1-a_{k})^{2}|/r)\times 100, \text{ or } (1-\sum_{k=1}^{r}|a_{kl}^{*}-(1-a_{k})|/r)\times 100, & \text{for Case 7} \\ (1-\sum_{k=1}^{r}|a_{kl}^{*}-(1-a_{k})^{\frac{1}{2}}|/r)\times 100, \text{ or } (1-\sum_{k=1}^{r}|a_{kl}^{*}-(1-a_{k})|/r)\times 100, & \text{for Case 8} \\ (1-\sum_{k=1}^{r}|a_{kl}^{*}-a_{k}|/r)\times 100, & \text{for Case 9} \\ (1-\sum_{k=1}^{r}|a_{kl}^{*}-slightly\ tilted\ of\ a_{k}|/r)\times 100, & \text{for Case 10} \end{cases}, (\%) \quad (13)$$

In Case 10 the given premise is $B^{*} = slightly\ tilted\ of\ B$, Conclusion $A^{*} = slightly\ tilted\ of\ A$.

**Definition 3.6.** *The reductive property criterion function* $RPCF_{FMT-FR}$ *for FMT are defined as formula (14).*

$$RPCF_{FMT-FR} = \frac{1}{S} \sum_{l=1}^{s} RPCF_{FMT-FR}^{l}, \quad (\%) \qquad (14)$$

The reductive property of fuzzy reasoning can be considered as *the reductive property of a fuzzy reasoning method or algorithm = average of (reductive property for FMP and reductive property for FMT)*.

**Definition 3.7.** *The criterion function for checking of the reductive property of fuzzy reasoning method is defined as arithmetic average value of* $RPCF_{FMP-FR}$ *and* $RPCF_{FMT-FR}$.

$$RPCF_{FR} = \tfrac{1}{2}(RPCF_{FMP-FR} + RPCF_{FMT-FR}), \quad (\%) \qquad (15)$$

In formula (13)−(15), indexes are the same as formula (10)−(11). According to above two definition, when the reduction property criterion function $RPCF_{FMP-FR}=100(\%)$ and $RPCF_{FMT-FR}=100(\%)$, then the reductive property of fuzzy reasoning method (or algorithm) is completely satisfied. This means that the given antecedent vector is equal to fuzzy reasoning result vector, that is, $b_k^* = b_k$, $k=1, 2, ..., r$, i.e., $B^* = B$, (resp. $a_k^* = a_k$, $k=1, 2, ..., r$, i.e., $A^* = A$), for FMP (resp. FMT). In other words, the larger $RPCF_{FMP-FR}$ (resp. $RPCF_{FMT-FR}$) is, the more the result of fuzzy reasoning satisfies the reductive property, and the smaller $RPCF_{FMP-FR}$ (resp. $RPCF_{FMT-FR}$) is, the less it satisfies. At worst, when criterion function $RPCF_{FMP-FR}=0(\%)$ and $RPCF_{FMT-FR}=0(\%)$, then the fuzzy reasoning method does not completely satisfy. Therefore the reductive property criterion function about every fuzzy reasoning method in fuzzy systems satisfies $0 \le RPCF_{FMP-FR} \le 100$, $0 \le RPCF_{FMT-FR} \le 100$ for FMP and FMT, respectively. This definitions differ largely from the several previous ones [2, 4, 8]. Therefore according to our definition method the fuzzy reasoning result can be more correctly evaluated, and effectively used in a lot of the practical problems.

Now let us discuss checking for the reductive property of fuzzy reasoning method.

**Example 3.1.** Assume that the fuzzy sets of the rule are given as $A(x) = [1, 0.3, 0, 0, 0]$, $B(y) = [0, 0, 0, 0.3, 1]$, and the given premise for FMP $[medium] = [0, 0.3, 1, 0.3, 0]$ $A^*(x) = A(x) = [1, 0.3, 0, 0, 0]$, the premise for FMT $B^*(y) = B(y) = [0, 0, 0, 0.3, 1]$, then the new conclusion reasoning result by any fuzzy reasoning method (for instance $WW$) is obtained as $B^*(y) \ne B(y) = [0, 0, 0.1, 0.4, 1]$ for FMP, $A^*(x) \ne A(x) = [1, 0.7, 0.4, 0.1, 0]$ for FMT, respectively. At this time according to our new method, the criterion function is calculated as follows.

$$RPCF_{FMP-WW-I} = (1 - \sum_{k=1}^{5} |b_k^* - b_k|/r) \times 100 = (1 - (|0-0| + |0-0| + |0.1-0| + |0.4-0.3| + |1-1|)/5) \times 100 =$$
$$= (1 - (0 + 0 + 0.1 + 0.1 + 0)/5) \times 100 = (1 - 0.2/5) \times 100 = (1 - 0.04) \times 100 = 96(\%)$$

$$RPCF_{FMT-WW-I} = (1 - \sum_{k=1}^{5} |a_{kl}^* - (1-a_{kl})|/5) \times 100 =$$
$$= (1 - (|1 - (1-0.3)| + |0.7 - (1-0)| + |0.4 - (1-0)| + |0.1 - (1-0)| + |0 - (1-0)|)/5) \times 100 =$$
$$= (1 - (|1-0.7| + |0.7-1| + |0.4-1| + |0.1-1| + |0-1|)/5) \times 100 =$$
$$= (1 - (0.3 + 0.3 + 0.6 + 0.9 + 1)/5) \times 100 = (1 - 3.1/5) \times 100 = (1 - 0.62) \times 100 = 38(\%)$$

$$RPCF_{WW} = \tfrac{1}{2}(RPCF_{FMP-WW-I} + RPCF_{FMT-WW-I}) = \tfrac{1}{2}(RPCF_{FMP-WW} + RPCF_{FMT-WW}) = \tfrac{1}{2}(96 + 38) = 67(\%)$$

Consequently the reductive property of a fuzzy reasoning method $WW$ does satisfy as 67(%). But according to [2], since $B^*(y) \ne B(y)$ for FMP and $A^*(x) \ne A(x)$ for FMT, the reductive property of a fuzzy reasoning method $WW$ does not satisfy as 0(%), thus their evaluation is strict and not right.

**Example 3.2.** Let us consider the reductive property of Example 1 and 2 in [7]. The fuzzy sets of the rule are as $[small] = [1, 0.3, 0, 0, 0]$, $[large] = [0, 0, 0, 0.3, 1]$ and the premise $[medium] = [0, 0.3, 1, 0.3, 0]$. The conclusions by CRI are obtained as $B^*(y) = [1, 1, 1, 1, 1]$ for FMP, and $A^*(x) = [0.3, 0.7, 1, 1, 1]$ for FMT. According to our Definition 3.3, 3.6, and 3.7, the reductive property of a fuzzy reasoning method $WW$ is as follows.

$$RPCF_{FMP-FR-WW} = (1 - \frac{2 \times E(B^*, B) \times e(A^*, A)}{E(B^*, B) + e(A^*, A)}) \times 100 = (1 - \frac{2 \times (\sum_{k=1}^{5} |b_k^* - b_k|/5) \times (\sum_{k=1}^{5} |a_k^* - a_k|/5)}{(\sum_{k=1}^{5} |b_k^* - b_k|/5) + (\sum_{k=1}^{5} |a_k^* - a_k|/5)}) \times 100 = 43.27(\%)$$

$$RPCF_{FMT-FR-WW} = (1 - \frac{2 \times E(A^*, A) \times e(B^*, B)}{E(A^*, A) + e(B^*, B)}) \times 100 = (1 - \frac{2 \times (\sum_{k=1}^{5}|a_k^* - (1-a_k)|/5) \times (\sum_{k=1}^{5}|b_k^* - (1-b_k)|/5)}{(\sum_{k=1}^{5}|a_k^* - (1-a_k)|/5) + (\sum_{k=1}^{5}|b_k^* - (1-b_k)|/5)}) \times 100 = 89.5(\%)$$

$$RPCF_{WW} = \tfrac{1}{2}(RPCF_{FMP-WW-1} + RPCF_{FMT-WW-1}) = \tfrac{1}{2}(RPCF_{FMP-WW} + RPCF_{FMT-WW}) = \tfrac{1}{2}(43.27 + 89.5) = 66.385(\%).$$

Now let us consider the deference of ours and [16]'s checking method. [16]'s checking method is very strict and has some weakness. [16]'s weakness is as follows. For the same Case such Example 3.1, according to [16]'s checking method, the fuzzy reasoning result are $B^*(y) = [0, 0, 0.1, 0.4, 1] \neq B(y) = [0, 0, 0, 0.3, 1]$ for FMP, and $A^*(x) = [1, 0.7, 0.4, 0.1, 0] \neq A(x) = [1, 0.3, 0, 0, 0]$ for FMT, respectively. Then the reductive property of the fuzzy reasoning method *WW* does not satisfy by the criterion of Table 1, that is, it is not flexible and soft. For this evaluation, considering by [16]'s viewpoint, it does satisfy as 0(%) or does not satisfy as 100(%), vice versa. So in order to overcome [16]'s weakness, we generalized and extended the criterion for FMP and FMT shown in Table 1 according to [16]. Frankly speaking, even though conclusion results are obtained as $B^*(y) \neq B(y)$ for FMP, and $A^*(x) \neq A(x)$ for FMT, respectively, our checking method can discuss the degree of coincidence between the given premise and the antecedent of fuzzy rule. In other words our proposed criterion function (15) tries to calculate the percentage degree of coincidence between the consequent $B(y)$ (resp. $A(x)$) of fuzzy rule and the conclusion $B^*(y)$ (resp. $A^*(x)$) of the reasoning, and then calculate the average of 2 percentage degrees of coincidence for FMP and FMT. Thereby the higher the degree of coincidence between $B(y)$ (resp. $A(x)$) and $B^*(y)$ (resp. $A^*(x)$) is, the better the reductive property of FMP (resp. FMT) is. Therefore as shown in Example 3.1 and 3.2, our new checking method of the reductive property is soft and well in accordance with general human understanding and practical problems than [16]'s one.

### 3. 3. New Fuzzy Reasoning Method For FMP

In this subsection we define several concepts and formulate new FMP-DM method based on distance measure.

According to the paper [24], distance measure is as follows. Let $F_0(R)$ be all continuous fuzzy subsets of $R$ whose $\alpha$-cuts are always bounded intervals. These will be called fuzzy numbers and are the fuzzy sets most widely used in practical applications. We need to be able to compute the distance between any fuzzy set $A$ and $B$ in $F_0(R)$. We know how to find the distance between two real numbers $x, y$. The distance is $|x - y| = DM(x, y)$. We also know how to find the distance between two points in $R_2$. The function $DM(x, y)$ used to compute distance is called a distance measure (DM). The basic properties of DM, i.e., $DM(x, y)$ for every $x, y$ in real space $R$ are:

① . $DM(x, y) \geq 0$; i.e., distance is not negative;
② . $DM(x, y) = DM(y, x)$; i.e., distance is symmetric;
③ . $DM(x, y) = 0$; if and only if $x = y$; i.e., we get zero distance only when $x = y$.
④ . $DM(x, y) \geq DM(x, z) + DM(z, y)$; i.e., it is shorter to go directly from $x$ to $y$ instead of first going to intermediate point $z$.

**Definition 3.8.** *Let the antecedent and the given premises $A$, $A_l^*$ for FMP be their discrete vector $A = [a_1, a_2, ..., a_k, ..., a_r]$, $A_l^* = [a_{1l}^*, a_{2l}^*, ..., a_{kl}^*, ..., a_{rl}^*]$, $(k = 1, 2, ..., r)$, respectively, where $a_k, a_{kl}^*$ are individual element of $A$, $A_l^*$, which are membership values in its fuzzy set, respectively. For FMP the individual elements $\alpha_{kl}$ of difference vector $\alpha_l = [\alpha_{1l}, \alpha_{2l}, ..., \alpha_{kl}, ..., \alpha_{rl}]$ are defined as follows, respectively.*

$$\alpha_{kl} = a_{kl}^* - a_k, \text{ for FMP} \tag{16}$$

**Definition 3.9.** *Let a discrete sign vector be $p_l = [p_{1l}, p_{2l}, ..., p_{kl}, ..., p_{rl}], (l = 1, 2, ..., s)$. Then element $p_{kl}$ of the sign vector is defined by two ways, i.e., $P(+1, 0, -1)$ form and $P(+1, -1)$ form, for FMP, as following formula, respectively.*

$$P(+1,0,-1)\,form \qquad P_{kl} = sign(\alpha_{kl}) = \begin{cases} +1, & \alpha_{kl} > 0 \\ 0, & \alpha_{kl} = 0 \\ -1, & \alpha_{kl} < 0 \end{cases}, \text{ for FMP} \qquad (17)$$

$$P(+1,-1)\,form \qquad P_{kl} = sign(\alpha_{kl}) = \begin{cases} +1, & \alpha_{kl} \geq 0 \\ -1, & \alpha_{kl} < 0 \end{cases}, \text{ for FMP} \qquad (18)$$

**Definition 3.10.** (See [10]) *For FMP, the distance measure $DM(A_l^*, A)$ between the antecedent fuzzy set $A$ and the given premise $A_l^*$ by using Euclidian distance measure is defined as follows.*

$$DM(A_l^*, A) = \left[ \sum_{k=1}^{r} \left[ a_{kl}^* - a_k \right]^2 / r \right]^{1/2}, \text{ for FMP} \qquad (19)$$

**Definition 3.11.** *The quasi-fuzzy reasoning result $\widetilde{B}_l$ for FMP can be defined as follows.*

$$\widetilde{B}_l = \begin{cases} B + DM(A_l^*, A) \times P_l, & \text{if Case 1, 2, and 3} \\ 1 - B + DM(A_l^*, A) \times P_l, & \text{if Case 4}, \qquad \text{for FMP} \\ \text{slightly tilted of } B + DM(A_l^*, A) \times P_l, & \text{if Case 5} \end{cases} \qquad (20)$$

**Definition 3.12.** *The maximum $\xi_l$ and minimum $\eta_l$ of the quasi-fuzzy reasoning result $\widetilde{B}_l$ are defined as follows, respectively.*

$$\xi_l = \max_{1 \leq k \leq r} \widetilde{B}_l, \qquad \eta_l = \min_{1 \leq k \leq r} \widetilde{B}_l \qquad \text{for FMP} \qquad (21)$$

**Definition 3.13.** *The fuzzy reasoning conclusion result for solving fuzzy modus ponens problem based on DM can be defined as formula (22), in this paper.*

$$(\text{FMP-DM}) \qquad B_l^* = \begin{cases} \dfrac{\widetilde{B}_l - \eta_l}{\xi_l - \eta_l}, & A_l^* \cap A \neq \Phi \\ 0, & A_l^* \cap A = \Phi \end{cases} \qquad (22)$$

Where $l = 1, 2, \cdots, s$ is index of the given premises $A_l^*$ for FMP, that is, $B_l^*$ is fuzzy reasoning conclusion by the *lth* given premise $A_l^*$ for FMP. And $\Phi \in F(X)$ is an empty set, $X$ is universe of discourse, and $x \in X$, $A \in F(X)$. Here, $A_l^*$, $\widetilde{A}_l$ and $A$ are the fuzzy sets in $F(X)$. The formula (22) is an standardization expression of the quasi-fuzzy reasoning result $\widetilde{B}_l$ for FMP.

The proposed method expressed by formula (22) is called distance measure method for the FMP with single input single output fuzzy system in this paper, for short FMP-DM. When combined $B_l^*$ and $A_l^*$, the fuzzy reasoning conclusion $B^*$ for FMP-DM can be described as follows.

$$B^* = \bigcup_{l=1}^{s} B_l^*, \qquad \text{for FMP-DM} \qquad (23)$$

Where $\cup$ is not max, but means the union of individual fuzzy sets obtained by fuzzy reasoning for FMP. Consequently, as defined in subsection 3.2, the criterion function $RPCF_{FR}$ for checking of the reductive property of fuzzy reasoning method is reflecting the degree of consistency between consequent $B$ and conclusion $B^*$ by formula (23), which is based on the degree of consistency between the antecedent $A$ and the given premise $A^*$ for FMP. Therefore it can be reasonable to consider the degree of consistency between conclusion $B^*$ and consequent $B$ to evaluate the reductive property (or reducibility) of FMP with considering the degree of consistency between the antecedent $A$ and the given premise $A^*$. For classical 2-valued logic, general modus ponens may be interpreted as if 《 if $x$ is $A$ then $y$ is $B$ 》 and $A^* = A$ then $B^* = B$. According to fuzzy logic, we hope to provide logical analysis for fuzzy modus ponens. Based on distance measure, FMP solution can be interpreted as if 《 if $x$ is $A$ then $y$ is $B$ 》 and 《 $A^*$ is closer to $A$ 》 then 《 $B^*$ is closer to $B$ 》. From the logical analysis of FMP solution, we can find that the conclusion $B^*$ not only relates to $A^*$ and 《 if $x$ is $A$ then $y$ is $B$ 》, but also relates to the distance measure of $A^*$ and $A$. How to select $DM(B^*, B)$ to make the conclusion of fuzzy reasoning more reasonable? We hope that $DM(B^*, B)$ is equal to $DM(A^*, A)$. And this property is proper with respect to fuzzy reasoning. Our aim is to search the fuzzy sets $B^*$ such that the distance measure $DM(B^*, B)$ should be fully supported by distance measure $DM(A^*, A)$. That is, following formula should be satisfied.

$$DM(B^*, B) = DM(A^*, A), \text{ for FMP} \tag{24}$$

There are a lot of fuzzy subsets on $Y$ that satisfy the formula (3). We hope the fuzzy subset as the conclusion of fuzzy reasoning satisfying the reductive property to be selected as soon as possible.

***Principle for solving of FMP-DM Problem.*** *The FMP-DM conclusion $B^*$ of formula (1) for a distance measure is the fuzzy subset of Y satisfying formula (24).*

According to this principle, FMP-DM method is as follows.

**Theorem 3.1.** *Assume that distance measure is Euclidean metric, then the FMP-DM solution of the formula (1) satisfying the formula (24) is described as follows.*

$$B^* = \begin{cases} f(B + DM(B^*, B)), & \text{if Case 1, 2, and 3} \\ f(1 - B + DM(B^*, B)), & \text{if Case 4} \\ f(\text{slightly tilted of } B + DM(B^*, B)), & \text{if Case 5} \end{cases}, \text{ for FMP} \tag{25}$$

*, where $f$ is standardization operator. Therefore there is no information loss of the fuzzy reasoning processing by $f$.*

**Proof.**

(i) Let's consider for Case 1, Case 2, and Case 3. For FMP-DM it is evident that if $A_l^* \cap A = \Phi$ then $B^* = 0$. When $A_l^* \cap A \neq \Phi$ then the fuzzy reasoning conclusion for FMP is obtained as follows.

$$B^* = \bigcup_{l=1}^{s} B_l^* = B_1^* \cup B_2^* \cup \cdots \cup B_l^* \cup \cdots \cup B_s^* =$$
$$= (\widetilde{B}_1 - \eta_1)/(\xi_1 - \eta_1) \cup (\widetilde{B}_2 - \eta_2)/(\xi_2 - \eta_2) \cup \cdots \cup (\widetilde{B}_l - \eta_l)/(\xi_l - \eta_l) \cup \cdots \cup (\widetilde{B}_s - \eta_s)/(\xi_s - \eta_s)$$
$$= (B_1 + DM(A_1^*, A) \times P_1 - \eta_1)/(\xi_1 - \eta_1) \cup (B_2 + DM(A_2^*, A) \times P_2 - \eta_2)/(\xi_2 - \eta_2) \cup \cdots,$$
$$\cup (B_l + DM(A_l^*, A) \times P_l - \eta_l)/(\xi_l - \eta_l) \cup \cdots \cup (B_s + DM(A_s^*, A) \times P_s - \eta_s)/(\xi_s - \eta_s)$$
$$= (B_1 \cup B_2 \cup \cdots \cup B_l \cup \cdots \cup B_s) + (DM(A_1^*, A) \times P_1 - \eta_1)/(\xi_1 - \eta_1) \cup DM(A_2^*, A) \times P_2 - \eta_2)/(\xi_2 - \eta_2)$$
$$\cup \cdots \cup DM(A_l^*, A) \times P_l - \eta_l)/(\xi_l - \eta_l) \cup \cdots \cup DM(A_s^*, A) \times P_s - \eta_s)/(\xi_s - \eta_s))$$
$$= \bigcup_{l=1}^{s} B_l + \bigcup_{l=1}^{s} DM(A_l^*, A) \times P_l - \eta_l)/(\xi_l - \eta_l$$
$$= \bigcup_{l=1}^{s} B_l + (\bigcup_{l=1}^{s} f(DM(A_l^*, A)))$$
$$= \bigcup_{l=1}^{s} B_l + f(\bigcup_{l=1}^{s} DM(A_l^*, A))$$
$$= f(B + DM(A^*, A))$$
$$= f(B + DM(B^*, B))$$

(ii) Let's consider for Case 4. The proof of (ii) is similar to (i), so it is abbreviated.

(iii) Let's consider for Case 5. The proof of (iii) is also similar to (i), thus it is also abbreviated.

Thus we have proved that fuzzy reasoning conclusion $B^*$ for FMP-DM obtained by the formula (25) satisfies the formula (24). The information loss is guaranteed by maximum $\xi_l$ and minimum $\eta_l$ of quasi-reasoning result $\widetilde{B}_l$ in above formula for FMP. □

Below an example for FMP-DM is shown.

**Example 3.3.** Let us consider for FMP-DM in Class 1. According to distance measure for FMP, FMP-DM can be obtained as the formula (22). For the antecedent $A = [1, 0.3, 0, 0, 0]$, the consequent $B = [0, 0, 0, 0.3, 1]$, for four Cases, fuzzy reasoning results are as follows. In Case 1, the given premise is $A^* = A$, distance measure $DM$ is calculated as $DM = 0$, quasi-reasoning result $\widetilde{B}$ for FMP-DM is calculated as $\widetilde{B} = [0, 0, 0, 0.3, 1]$, since $B^* = \widetilde{B} = [0, 0, 0, 0.3, 1] = B$, therefore the reductive property is 100(%). In Case 2, the given premise is $A^* = A^2 = [1, 0.09, 0, 0, 0]$, and the proposed distance measure DM is calculated as following formula. $DM = [((1-1)^2 + (0.09-0.3)^2 + 1^2 + 0^2 \times 3)/5]^{1/2} = 0.094$. And then quasi-reasoning result for FMP-DM is calculated as $\widetilde{B} = [0.094, -0.094, 0.094, 0.394, 1.094]$, where maximum and minimum of $\widetilde{B}$ are $\xi = 1.094$, and $\eta = -0.094$. Therefore the fuzzy reasoning result in Case 2 is calculated as $B^* = [0.0859, 0, 0.0859, 0.36, 1]$. When compared with the consequent $B^2 = [0, 0, 0, 0.09, 1]$ of the fuzzy rule, the reductive property for FMP-DM in Case 2 is calculated as 91.16(%). In Case 3, the given premise is $A^* = A^{1/2} = [1, 0.55, 0, 0, 0]$, $DM$ for FMP is calculated as $DM = [(0.55-0.3)^2/5]^{1/2} = 0.112$, the quasi-reasoning result for FMP-DM is calculated as quasi-reasoning result $\widetilde{B} = [0.112, 0.112, 0.112, 0.412, 1.112]$, where maximum and minimum of the quasi-reasoning result $\widetilde{B}$ are $\xi = 1.112$, $\eta = 0.112$, since $B^* = [0, 0.111, 0, 0.3, 1]$, $B^{1/2} = [0, 0, 0, 0.55, 1]$. Therefore the reductive property in Case 3 is calculated as 92.83 (%). And in Case 4, the given premise is $A^* = 1 - A = [0, 0.7, 1, 1, 1]$, DM for FMP is calculated as $DM = [(1^2 + 0.4^2 + 1^2 + 1^2 + 1^2)/5]^{1/2} = 0.912$, since the quasi-reasoning result for FMP-DM is calculated as $\widetilde{B} = [-0.912, 0.912, 0.912, 1.212, 1.912]$, where maximum and minimum of the quasi-reasoning result $\widetilde{B}$ are $\xi = 1.912$, $\eta = -0.912$, therefore, for FMP-DM, fuzzy reasoning result is obtained as $B^* = [0, 1, 1, 0.84, 0.45]$. When compared with the consequent fuzzy set of the fuzzy rule $\overline{B} = [1, 1, 1, 0.7, 0]$, the reductive property for FMP-DM in Case 4 is calculated as 68.25(%). Thus total reductive property criterion function value for FMP-DM presented in this paper is obtained as $RPCF_{FMP-DM}$ = (100+91.16+92.83+68.25)/4=87.7285 (%).

### 3. 4. New Fuzzy Reasoning Method For FMT

In this subsection we define several concepts and formulate new FMT-DM method based on distance measure [24].

**Definition 3.14.** Let the antecedent $B$ and the given premises $B_l^*$ for FMT be their discrete vector $B = [b_1, b_2, ..., b_k, ..., b_r]$, and $B_l^* = [b_{1l}^*, b_{2l}^*, ..., b_{kl}^*, ..., b_{rl}^*]$, $(k = 1, 2, ..., r)$, respectively. Where $b_k$, and $b_{kl}^*$ are individual elements of $B$ and $B_l^*$, which are membership values in its fuzzy set, respectively. For FMT the individual elements $\beta_{kl}$ of difference vector $\beta_l = [\beta_{1l}, \beta_{2l}, ..., \beta_{kl}, ..., \beta_{rl}]$ are defined as follows, respectively.

$$\beta_{kl} = b_{kl}^* - b_k, \text{ for FMT} \tag{26}$$

**Definition 3.15.** Let a discrete sign vector be $p_l = [p_{1l}, p_{2l}, ..., p_{kl}, ..., p_{rl}]$, $(l = 1, 2, ..., s)$. Then element $p_{kl}$ of the sign vector is defined by two ways, i.e., $P(+1, 0, -1)$ form and $P(+1, -1)$ form, for FMT, as following formulas, respectively.

$$P(+1, 0, -1) form \quad P_{kl} = sign(\beta_{kl}) = \begin{cases} +1, & \beta_{kl} > 0 \\ 0, & \beta_{kl} = 0 \\ -1, & \beta_{kl} < 0 \end{cases}, \text{ for FMT} \tag{27}$$

$$P(+1,-1) \text{ form} \qquad P_{kl} = sign(\beta_{kl}) = \begin{cases} +1, & \beta_{kl} \geq 0 \\ -1, & \beta_{kl} < 0 \end{cases}, \text{ for FMT} \qquad (28)$$

**Definition 3.16.** (See [10]) *For FMT, the distance measure* $DM(B_l^*, B)$ *between the antecedent fuzzy set* $B$ *and the given premise* $B_l^*$ *by using Euclidian distance measure is defined as follows, according to the paper [10].*

$$DM(B_l^*, B) = \left[ \sum_{k=1}^{r} \left[ b_{kl}^* - b_k \right]^2 / r \right]^{1/2}, \text{ for FMT} \qquad (29)$$

**Definition 3.17.** *The quasi-fuzzy reasoning result* $\widetilde{A}_l$ *for FMT can be defined as follows.*

$$\widetilde{A}_l = \begin{cases} 1 - A + DM(B_l^*, B) \times P_l, & \text{if Case 6, 7, and 8} \\ A + DM(B_l^*, B) \times P_l, & \text{if Case 9}, \text{ for FMT} \\ \text{slightly tilted of } A + DM(B_l^*, B) \times P_l, & \text{if Case 10} \end{cases} \qquad (30)$$

**Definition 3.18.** *The maximum* $\xi_l$ *and minimum* $\eta_l$ *of the quasi-fuzzy reasoning result* $\widetilde{A}_l$, $(l = 1, 2, \cdots, s)$ *are defined as follows, respectively.*

$$\xi_l = \max_{1 \leq k \leq r} \widetilde{A}_l, \qquad \eta_l = \min_{1 \leq k \leq r} \widetilde{A}_l \qquad \text{for FMT} \qquad (31)$$

**Definition 3.13.** *The fuzzy reasoning results for solving of fuzzy modus tollens based on DM is defined as formula (32).*

$$\text{(FMT-DM)} \qquad A_l^* = \begin{cases} \dfrac{\widetilde{A}_l - \eta_l}{\xi_l - \eta_l}, & B_l^* \cap B \neq \Phi \\ 0, & B_l^* \cap B = \Phi \end{cases} \qquad (32)$$

Where $l = 1, 2, \cdots, s$ is index of the given premises $B_l^*$ for FMT, that is, $A_l^*$ is fuzzy reasoning conclusion by the *l*th given premise $B_l^*$ for FMT. And $\Phi \in F(Y)$ is an empty set for FMT, also $Y$ is universe of discourse, and $y \in Y$, $B \in F(Y)$. Here, $B_l^*$, $\widetilde{B}_l$, and $B$ are the fuzzy sets in $F(Y)$. The formula (32) is a standardization expression of the quasi-fuzzy reasoning result $\widetilde{A}_l$ for FMT. The proposed method expressed by formula (32) is called distance measure method of fuzzy reasoning for FMT with single input single output fuzzy system in this paper, for short FMT-DM. When combined $B_l^*$ and $A_l^*$, the fuzzy reasoning conclusion $A^*$ for FMT can be described as follows.

$$A^* = \bigcup_{l=1}^{s} A_l^*, \quad \text{for FMT} \qquad (33)$$

Where $\cup$ is not max, means the union of individual fuzzy sets obtained by fuzzy reasoning for FMT.

Consequently, as defined in subsection 3.1, the criterion function $RPCF_{FR}$ for checking of the reductive property of fuzzy reasoning method is reflecting the degree of consistency between consequent $A$ and conclusion $A^*$ by formula (32), which is based on the degree of consistency between the antecedent $B$ and the given premise $B^*$ for FMT. Therefore it can be reasonable to consider the degree of consistency between the fuzzy reasoning conclusion $A^*$ and consequent $A$ to evaluate the reductive property (or reducibility) of FMT with considering the consistency between the antecedent $B$ and the given premise $B^*$. For classical 2-valued logic, general modus ponens may be interpreted as if 《 *if x is A then y is B* 》 and $A^* = A$ then $B^* = B$. According to fuzzy logic, we hope to provide logical analysis for fuzzy modus ponens. Based on distance measure, FMP solution can be interpreted as if 《 *if x is A then y is B* 》 and 《 $A^*$ *is closer to A* 》 then 《 $B^*$ *is closer to B* 》. From the logical analysis of FMP solution, we can find that the conclusion $B^*$ not only relates to $A^*$ and 《 *if x is A then y is B* 》, but also relates to the distance measure of $A^*$ and $A$. How to select $DM(B^*, B)$ to make the conclusion of fuzzy reasoning more reasonable? We hope that $DM(B^*, B)$ is equal to $DM(A^*, A)$. And this property is proper with respect to fuzzy reasoning. Our aim is to search the fuzzy sets $B^*$ such that the distance measure $DM(B^*, B)$ should be fully supported by distance measure $DM(A^*, A)$.

Let us consider FMT-DM. Here our aim is to search the fuzzy sets $A^*$ such that the distance measure $DM(A^*, \overline{A})$ obtained by the fuzzy reasoning conclusion and the consequent should be fully supported by distance measure $DM(B^*, \overline{B})$ obtained by the given premise and the antecedent. That is, following formula for FMT should be satisfied.

$$DM(A^*, \overline{A}) = DM(B^*, \overline{B}) \quad , \text{for FMT} \tag{34}$$

There are a lot of fuzzy subsets on *X* that satisfy the formula (3). We try to select the fuzzy subset as the conclusion of fuzzy reasoning satisfying the reductive property.

***Principle for solving of FMT-DM Problem.*** *The FMT-DM conclusion $A^*$ of the formula (3) for a distance measure is the fuzzy subset of X satisfying the formula (34).*

According to this principle, FMT-DM method is as follows.

**Theorem 3.2.** *Assume that distance measure is Euclidean metric, then the FMT-DM solution of the formula (3) satisfying the formula (34) is expressed as follows.*

$$A^* = \begin{cases} f(\overline{A} + DM(A^*, A)), & \text{if Case 6, 7, and 8} \\ f(A + DM(A^*, A)), & \text{if Case 9}, \text{ for FMT} \\ f(\text{slightly tilted of } A + DM(A^*, A)) & \text{if Case 10} \end{cases} \tag{35}$$

*Where $f$ is standardization operator. Hereby there is no information loss of the fuzzy reasoning processing by $f$.*

**Proof.** For the proof we consider 3 cases of the formula (35) for FMT of proposed DMM.

(i) Let's consider for Case 6, Case 7, and Case 8 in formula (35).

For FMT-DM it is evident that if $B_l^* \cap \overline{B} = \Phi$ then $A^* = 0$. Now, when $B_l^* \cap \overline{B} \neq \Phi$ then the fuzzy reasoning conclusion is obtained as follows.

$$A^* = \bigcup_{l=1}^{s} A_l^* = A_1^* \cup A_2^* \cup \cdots A_l^* \cup \cdots A_s^* =$$

$$= (\widetilde{A}_1 - \eta_1)/(\xi_1 - \eta_1) \cup (\widetilde{A}_2 - \eta_2)/(\xi_2 - \eta_2) \cup \cdots \cup (\widetilde{A}_l - \eta_l)/(\xi_l - \eta_l) \cup \cdots \cup (\widetilde{A}_s - \eta_s)/(\xi_s - \eta_s)$$

$$= (\overline{A}_1 + DM(A_1^*, \overline{A}) \times P_1 - \eta_1)/(\xi_1 - \eta_1) \cup (\overline{A}_2 + DM(A_2^*, \overline{A}) \times P_2 - \eta_2)/(\xi_2 - \eta_2) \cup \cdots,$$

$$\cup (\overline{A}_l + DM(A_l^*, \overline{A}) \times P_l - \eta_l)/(\xi_l - \eta_l) \cup \cdots \cup (\overline{A}_s + DM(A_s^*, \overline{A}) \times P_s - \eta_s)/(\xi_s - \eta_s)$$

$$= (\overline{A}_1 \cup \overline{A}_2 \cup \cdots \cup \overline{A}_l \cup \cdots \overline{A}_s) + (DM(B_1^*, \overline{B}) \times P_1 - \eta_1)/(\xi_1 - \eta_1) \cup DM(B_2^*, \overline{B}) \times P_2 - \eta_2)/(\xi_2 - \eta_2)$$

$$\cup \cdots \cup DM(B_l^*, \overline{B}) \times P_l - \eta_l)/(\xi_l - \eta_l) \cup \cdots \cup DM(B_s^*, \overline{B}) \times P_s - \eta_s)/(\xi_s - \eta_s))$$

$$= \bigcup_{l=1}^{s} \overline{A}_l + \bigcup_{l=1}^{s} DM(B_l^*, \overline{B}) \times P_l - \eta_l)/(\xi_l - \eta_l)$$

$$= \bigcup_{l=1}^{s} \overline{A}_l + (\bigcup_{l=1}^{s} f(DM(B_l^*, \overline{B})))$$

$$= \bigcup_{l=1}^{s} \overline{A}_l + f(\bigcup_{l=1}^{s} DM(B_l^*, \overline{B}))$$

$$= f(\overline{A} + DM(B_l^*, \overline{B}))$$

$$= f(\overline{A} + DM(A^*, \overline{A}))$$

(ii) Let's consider for Case 9 in formula (35). The proof (ii) is similar to (i).

(iii) Let's consider for Case 10 in formula (35). The proof (iii) is also similar to (i).

Thus we have proved that fuzzy reasoning conclusion $A^*$ for FMT-DM obtained by the formula (35) satisfies the formula (34). The information loss is guaranteed by maximum $\xi_l$ and minimum $\eta_l$ of quasi-reasoning result $\widetilde{A}_l$ in above formula for FMT. □

## 4. Checking of QIP, CRI, TIP, AARS, and Our Proposed DMM

In this section as mentioned in abstract we check Example of "The Quintuple Implication Principle of fuzzy reasoning" (QIP) presented in "Information Sciences", 297 (2015) 202-215 with respect to "Comparison of fuzzy reasoning methods" in "Fuzzy Sets and Systems" 8 (1982). Since QIP (2015) presented by Baokui Zhou, Genqi Xu and Sanjiang Li is less than CRI(1973) presented by Lotfi A. Zadeh, with respect to the reductive property, the illustrative checking in this section has very important significance.

### 4.1. Checking of FMP-QIP and FMT-QIP

In this subsection, we check the reductive property of FMP-QIP by using Implication of Łukasiewicz, Gödel, $R_0$ and Gougen. Proposition 5 and Theorem 2 in [33] is as follows.

**Proposition 4.1**. *(See Proposition 5 in [33]) The QIP method for FMP (FMT, resp.) is recoverable if A (B, resp.) is normal, where a fuzzy set F on universe W is normal if there exists $w \in W$ such that $F(w) = 1$.*

**Theorem 4.1.** *(See Theorem 2 in [33]) Suppose $\otimes$ is a left continuous t-norm and $\rightarrow$ its residual implication. Then the QIP solution of FMP and FMT is as follows:*

$$\text{(FMP-QIP)} \quad B^*(y) = \bigvee_{x \in U} (A^*(x) \otimes (A^*(x) \rightarrow A(x)) \otimes (A(x) \rightarrow B(y))) \tag{36}$$

$$\text{(FMT-QIP)} \quad A^*(x) = \bigvee_{y \in V} (A(x) \otimes (A(x) \rightarrow B(y)) \otimes (B(y) \rightarrow B^*(y))) \tag{37}$$

When the fuzzy sets of the rule are given as $A(x) = [1, 0.3, 0, 0, 0]$, $B(y) = [0, 0, 0, 0.3, 1]$, and the four premises $A^*(x) = A(x) = [1, 0.3, 0, 0, 0]$, $A^*(x) = A^2(x) = [1, 0.09, 0, 0, 0]$, $A^*(x) = A^{1/2}(x) = [1, 0.548, 0, 0, 0]$, and $A^*(x) = \overline{A(x)} = [0, 0.7, 1, 1, 1]$, the fuzzy reasoning results based on Quintuple Implication Principle for FMP (formula (13), i.e., formula (36) in [7]) are calculated as follows. For example, the fuzzy reasoning result of FMP-QIP-Łukasiewicz for Case 1 is as follows.

$$B^*(y) = \bigvee_{x \in U} (A(x) \otimes_L (A(x) \rightarrow_L A(x)) \otimes_L (A(x) \rightarrow_L B(y))) =$$

$$= \bigvee_{x \in U} \left( [1, 0.3, 0, 0, 0] \otimes_L \begin{bmatrix} 0 & 0 & 0 & 0.3 & 1 \\ 0.7 & 0.7 & 0.7 & 1 & 1 \\ 1 & 1 & 1 & 1 & 1 \\ 1 & 1 & 1 & 1 & 1 \\ 1 & 1 & 1 & 1 & 1 \end{bmatrix} \right) = [0, 0, 0, 0.3, 1] = B(y)(= [0, 0, 0, 0.3, 1])$$

The values of criterion function for FMP-QIP and FMT-QIP reductive property in Class 1 are as shown in Table 2.

**Table 2.** FMP-QIP and FMT-QIP reductive property in Class 1

| FMP-QIP Premise $A^*(x)$ | FMP-QIP Conclusion $B^*(y)$ and Reductive Property | | | |
|---|---|---|---|---|
| | FMP-QIP-Łukasiewicz | FMP-QIP-Gödel | FMP-QIP-$R_0$ | FMP-QIP-Gougen |
| [1, 0.3, 0, 0, 0] | [0, 0, 0, 0.3, 1] 100 % | [0, 0, 0, 0.3, 1] 100 % | [0, 0, 0, 0.3, 1] 100 % | [0, 0, 0, 0.3, 1] 100 % |
| [1, 0.09, 0, 0, 0] | [0, 0, 0, 0.3, 1] 95.8% | [0, 0, 0, 0.3, 1] 95.8% | [0, 0, 0, 0.3, 1] 95.8% | [0, 0, 0, 0.3, 1] 95.8% |
| [1, 0.548, 0, 0, 0] | [0, 0, 0, 0.3, 1] 95.1% | [0, 0, 0, 0.3, 1] 95.1% | [0, 0, 0, 0.3, 1] 95.1% | [0, 0, 0, 0.3, 1] 95.1% |
| [0, 0.7, 1, 1, 1] | [0, 0, 0, 0.3, 1] 26 % | [0, 0, 0, 0.3, 1] 26 % | [0, 0, 0, 0.3, 1] 26 % | [0, 0, 0, 0.3, 1] 26 % |
| $RPCF_{FMP-QIP-I}$ | 79.21 % | 79.21 % | 79.21 % | 79.21 % |
| FMT-QIP Premise $B^*(y)$ | FMT-QIP Conclusion $A^*(x)$ and Reductive Property | | | |
| | FMT-QIP-Łukasiewicz | FMT-QIP-Gödel | FMT-QIP-$R_0$ | FMT-QIP-Gougen |
| [1, 1, 1, 0.7, 0] | [0.3, 0.3, 0, 0, 0] 26 % | [0.3, 0.3, 0, 0, 0] 26 % | [0.3, 0.3, 0, 0, 0] 26 % | [0.3, 0.3, 0, 0, 0] 26 % |
| [1, 1, 1, 0.91, 0] | [0.3, 0.3, 0, 0, 0] 21.8 % | [0.3, 0.3, 0, 0, 0] 21.8 % | [0.3, 0.3, 0, 0, 0] 21.8 % | [0.3, 0.3, 0, 0, 0] 21.8 % |
| [1, 1, 1, 0.452, 0] | [0.3, 0.3, 0, 0, 0] 42.95% | [0.3, 0.3, 0, 0, 0] 42.95% | [0.3, 0.3, 0, 0, 0] 42.95 % | [0.3, 0.3, 0, 0, 0] 42.95 % |
| [0, 0, 0, 0.3, 1] | [1, 0.3, 0, 0, 0] 100 % | [1, 0.3, 0, 0, 0] 100 % | [1, 0.3, 0, 0, 0] 100 % | [1, 0.3, 0, 0, 0] 100 % |
| $RPCF_{FMT-QIP-I}$ | 47.69 % | 47.69 % | 47.69 % | 47.69 % |

Fuzzy reasoning results based on Quintuple Implication Principle in [7] are calculated by formula (36) and (37). The antecedent of the fuzzy rule is $A(x)=[1, 0.3, 0, 0, 0]$, consequent of the fuzzy rule $B(y)=[0, 0, 0, 0.3, 1]$. And the given premises are $B^*(y) = \text{not } B(y) = \overline{B(y)}=[1, 1, 1, 0.7, 0]$, $B^*(y) = \overline{B^2(y)}=[1, 1, 1, 0.91, 0]$, $B^*(y) = \overline{B^{1/2}(y)}=[1, 1, 1, 0.452, 0]$, and $B^*(y) = B(y) = [0, 0, 0, 0.3, 1]$. In Table 2 the reductive property is expressed as $RPCF_{FMP-QIP-I}$ and $RPCF_{FMP-QIP-I}$. From computational results shown in Table 2, we can see that the reductive property of FMP-QIP-Łukasiewicz, FMP-QIP-Gödel, FMP-QIP-$R_0$ and FMP-QIP-Gougen are all the same as 79.21%. FMT-QIP-I reductive properties are checked in Table 2. FMT-QIP by Łukasiewicz, Gödel, $R_0$, and Gougen satisfy all the reductive property, i.e., recovery in [7] as 47.69%, respectively.

**Proposition 4.2.** *The reductive property of FMP-QIP and FMT-QIP by Łukasiewicz, Gödel, $R_0$ and Gougen are all same with respect to [2], respectively, FMP-QIP is more than FMT-QIP.*

### 4.2. Checking of FMP and FMT by Zadeh's CRI

In this section we check the reductive property of FMP and FMT by Zadeh's CRI.
Definition 6 in [33] is as follows.

**Definition 4.1.** *(See Definition 6 in [33]) A method of FMP is said recoverable if $A^*(x) = A(x)$ implies $B^*(y) = B(y)$; similarly, an algorithm of FMT is recoverable if $B^*(y) = B(y)$ implies $A^*(x) = A(x)$.*

The most general forms of the CRI solutions of FMP and FMT are as follows.

$$\text{(FMP-CRI)} \quad B^*(y) = \bigvee_{x \in U} (A^*(x) \otimes (A(x) \to B(y))) \tag{38}$$

$$\text{(FMT-CRI)} \quad A^*(x) = \bigvee_{y \in V} (B^*(y)) \otimes (A(x) \to B(y))) \tag{39}$$

FMP-CRI and FMT-CRI reductive properties based on formula (38) and (39) in Class 1 are shown in Table 3.

**Table 3.** FMP-CRI and FMT-CRI reductive property in Class 1

| FMP-CRI Premise $A^*(x)$ | FMP-CRI-Conclusion $B^*(y)$ and Reductive Property | | | |
|---|---|---|---|---|
| | FMP-QIP-Łukasiewicz | FMP-QIP-Gödel | FMP-QIP-$R_0$ | FMP-QIP-Gougen |
| [1, 0.3, 0, 0, 0] | [0, 0, 0, 0.3, 1] 100 % | [0, 0, 0, 0.3, 1] 100 % | [0, 0, 0, 0.3, 1] 100 % | [0, 0, 0, 0.3, 1] 100 % |
| [1, 0.09, 0, 0, 0] | [0, 0, 0, 0.3, 1] 95.8 % | [0, 0, 0, 0.3, 1] 95.8 % | [0, 0, 0, 0.3, 1] 95.8 % | [0, 0, 0, 0.3, 1] 95.8 % |
| [1, 0.548, 0, 0, 0] | [0.248, 0.248, 0.248, 0.548, 1]   85.14 % | [0, 0, 0, 0.548, 1] 100 % | [0.548, 0.548, 0.548, 0.548, 1]   67.14 % | [0, 0, 0, 0.548, 1] 100 % |
| [0, 0.7, 1, 1, 1] | [1, 1, 1, 1, 1] 74 % | [1, 1, 1, 1, 1] 74 % | [1, 1, 1, 1, 1] 74 % | [1, 1, 1, 1, 1] 74 % |
| $RPCF_{FMP-CRI-I}$ | 88.73 % | 92.45 % | 84.23 % | 92.45 % |
| FMT-CRI Premise $B^*(y)$ | FMT-CRI-Conclusion $A^*(x)$ and Reductive Property | | | |
| | FMT-CRI-Łukasiewicz | FMT-CRI-Gödel | FMT-CRI-$R_0$ | FMT-CRI-Gougen |
| [1, 1, 1, 0.7, 0] | [1, 1, 1, 1, 1] 74 % | [1, 1, 1, 1, 1] 74 % | [1, 1, 1, 1, 1] 74 % | [1, 1, 1, 1, 1] 74 % |
| [1, 1, 1, 0.91, 0] | [1, 1, 1, 1, 1] 78.2 % | [1, 1, 1, 1, 1] 78.2 % | [1, 1, 1, 1, 1] 78.2 % | [1, 1, 1, 1, 1] 78.2 % |
| [1, 1, 1, 0.452, 0] | [1, 1, 1, 1, 1] 69.05 % | [1, 1, 1, 1, 1] 69.05 % | [1, 1, 1, 1, 1] 69.05 % | [1, 1, 1, 1, 1] 69.05 % |
| [0, 0, 0, 0.3, 1] | [1, 1, 1, 1, 1] 26 % | [1, 1, 1, 1, 1] 26 % | [1, 1, 1, 1, 1] 26 % | [1, 1, 1, 1, 1] 26 % |
| $RPCF_{FMT-CRI-I}$ | 61.81 % | 61.81 % | 61.81 % | 61.81 % |

When compared with Table 2, FMP-QIP reductive property (79.21 % for all 4 implications) is less than FMP-CRI (from 84.23 % for FMP-CRI-$R_0$, 88.73 % for FMP-CRI-Łukasiewicz to 92.45 % for FMP-CRI-Gödel and FMP-CRI-Gougen) one in Table 3. And FMT-QIP reductive property (47.69 % for all 4 implications) is less than FMT-CRI one (61.81 % for all 4 implications) in Table 3.

**Proposition 4.3.** *The reductive property of FMP-CRI and FMT-CRI by Łukasiewicz, Gödel, $R_0$ and Gougen are more than FMP-QIP and FMT-QIP with respect to [2], respectively.*

### 4.3. Reductive Property Checking of FMP and FMT by Wang's TIP

In Table 4 we show FMP-TIP and FMT-TIP reductive property, respectively. Theorem 1 in [33] is as follows.

**Theorem 4.2.** *(See Theorem 1 in [33]) Suppose $\otimes$ is a left continuous t-norm and $\rightarrow$ its residual. Then the TIP solution of FMP and FMT are as follows:*

$$\text{(FMP-TIP)} \quad B^*(y) = \bigvee_{x \in U} (A^*(x) \otimes (A(x) \rightarrow B(y))) \tag{40}$$

$$\text{(FMT-TIP)} \quad A^*(x) = \bigwedge_{y \in V} ((A(x) \rightarrow B(y)) \rightarrow B^*(y))) \tag{41}$$

The experimental result about the reductive property for FMP-TIP and FMT-TIP in Class 1 is shown in Table 4.

**Table 4.** FMP-TIP and FMT-TIP reductive property in Class 1

| FMP-TIP Premise $A^*(x)$ | FMP-TIP-Conclusion $B^*(y)$ and Reductive Property | | | |
|---|---|---|---|---|
| | FMP-TIP-Łukasiewicz | FMP-TIP-Gödel | FMP-TIP-$R_0$ | FMP-TIP-Gougen |
| [1, 0.3, 0, 0, 0] | [0, 0, 0, 0.3, 1] 100 % | [0, 0, 0, 0.3, 1] 100 % | [0, 0, 0, 0.3, 1] 100 % | [0, 0, 0, 0.3, 1] 100 % |
| [1, 0.09, 0, 0, 0] | [0, 0, 0, 0.3, 1] 95.8 % | [0, 0, 0, 0.3, 1] 95.8 % | [0, 0, 0, 0.3, 1] 95.8 % | [0, 0, 0, 0.3, 1] 95.8 % |
| [1, 0.548, 0, 0, 0] | [0.248, 0.248, 0.248, 0.548, 1] 85.14 % | [0, 0, 0, 0.548, 1] 100 % | [0.548, 0.548, 0.548, 0.548, 1] 67.14 % | [0, 0, 0, 0.548, 1]100% |
| [0, 0.7, 1, 1, 1] | [1, 1, 1, 1, 1] 74 % | [1, 1, 1, 1, 1] 74 % | [1, 1, 1, 1, 1] 74 % | [1, 1, 1, 1, 1] 74 % |
| $RPCF_{FMP-TIP-I}$ | 88.73 % | 92.45 % | 84.23 % | 92.45 % |
| FMT-TIP Premise $B^*(y)$ | FMT-TIP-Conclusion $A^*(x)$ and Reductive Property | | | |
| | FMT-TIP-Łukasiewicz | FMT-TIP-Gödel | FMT-TIP-$R_0$ | FMT-TIP-Gougen |
| [1, 1, 1, 0.7, 0] | [0, 0, 0, 0, 0] 26 % | [0, 0, 0, 0, 0] 26 % | [0, 0, 0, 0, 0] 26 % | [0, 0, 0, 0, 0] 26 % |
| [1, 1, 1, 0.91, 0] | [0, 0, 0, 0, 0] 21.8 % | [0, 0, 0, 0, 0] 21.8 % | [0, 0, 0, 0, 0] 21.8 % | [0, 0, 0, 0, 0] 21.8 % |
| [1, 1, 1, 0.452, 0] | [0, 0, 0, 0, 0] 30.95 % | [0, 0, 0, 0, 0] 30.95 % | [0, 0, 0, 0, 0] 30.95 % | [0, 0, 0, 0, 0] 30.95 % |
| [0, 0, 0, 0.3, 1] | [1, 0.3, 0, 0, 0] 100 % | [1, 0.3, 0, 0, 0] 100 % | [1, 0.3, 0, 0, 0] 100 % | [1, 0.3, 0, 0, 0] 100 % |
| $RPCF_{FMT-TIP-I}$ | 44.69 % | 44.69 % | 44.69 % | 44.69 % |

When compared with Table 2, FMP-QIP reductive property (79.21 % for all 4 implications) is less than FMP-TIP (from 84.23 % for FMP-TIP-$R_0$ and 88.73 % for FMP-TIP-Łukasiewicz to 92.45 % for FMP-TIP-Gödel and FMP-TIP-Gougen) one in Table 4. When compared with Table 2, FMT-QIP reductive property (47.69 % for all 4 implications) is more than FMP-TIP one (44.69 % for all 4 implications) in Table 4.

**Proposition 4.4.** *The reductive property of FMP-TIP and FMT-TIP by Łukasiewicz, Gödel, $R_0$ and Gougen are more than FMP-QIP and FMT-QIP with respect to [2], respectively.*

### 4.4. Checking of FMP and FMT by Turksen and Zhong's AARS

In this section, we check the reductive property, i.e., the recovery of Example 1 and 2 in [7] by using [8].

Unlike CRI [31], in [21], a similarity-based fuzzy reasoning method, i.e., Turksen and Zhong's Approximate Analogical Reasoning Schema (AARS) was proposed. The AARS modifies the consequent based on the similarity (closeness) between the fact, i.e., the given premise $A^*$ and the antecedent $A$. If the degree of similarity measure is greater than the predefined threshold value, then the rule will be fired and the consequent is deduced by some modification techniques. In [8], one of distance measures (DM) for FMP is as follows.

$$DM = D_2(A^*, A) = \left[ \sum_{i=1}^{n} \left[ \mu_{A^*}(x_i) - \mu_A(x_i) \right]^2 / n \right]^{1/2} \tag{42}$$

According to [8], distance measures (DM) for FMT is as follows.

$$DM = D_2(B^*, B) = \left[ \sum_{i=1}^{n} \left[ \mu_{B^*}(y_i) - \mu_B(y_i) \right]^2 / n \right]^{1/2} \tag{43}$$

The similarity by distance measures (DM) is then defined as follows.

$$S_{AARS} = (1+DM)^{-1} \qquad (44)$$

If the rule will be fired, then the consequent is modified by a modification function which could appear in one of the two forms for FMP and FMT i.e. *more or less form* and, *fuzzy membership value reduction form*, for short, *reduction form*, according to [21], respectively:

(FMP-AARS-*more or less form*) $\qquad B^* = \min\{1, B/S_{AARS}\} \qquad (45)$

(FMT-AARS-*more or less form*) $\qquad A^* = \min\{1, A/S_{AARS}\} \qquad (46)$

(FMP-AARS-*reduction form*) $\qquad B^* = B \times S_{AARS} \qquad (47)$

(FMT-AARS-*reduction form*) $\qquad A^* = A \times S_{AARS} \qquad (48)$

First, fuzzy reasoning for FMP-AARS-*more or less form* is as follows. In Table 1, from the given premise $A^* = A = [small]$, since DM and similarity presented by Turksen and Zhong are calculated as follows.

$$D_2(A^*, A) = \left[\sum_{i=1}^{5}[[1, 0.3, 0, 0, 0] - [1, 0.3, 0, 0, 0]]^2 / 5\right]^{1/2} = 0, \text{ and } S_{AARS} = \frac{1}{1+0} = 1.$$

The conclusion FMP-AARS -*more or less form* (45) is calculated as follows.

$B^* = \min\{1, [1, 0.3, 0, 0, 0]\} = [1, 0.3, 0, 0, 0] = B$.

Here, we applied FMP-AARS-*more or less form* specified as the formula (45). When compared with the QIP-Łukasiewicz, QIP-Gödel, QIP-Gougen and QIP-R$_0$ solutions, FMP-AARS presented by Turksen and Zhong is not much closer to the statement that ''y is large''. Thus, FMP-AARS-*more or less form* is not in accordance with human thinking. The fuzzy reasoning results of AARS for FMP and FMT are shown in Table 5. Next, fuzzy reasoning result for FMT-AARS-*more or less form* is calculated as $A^* = [1, 0.574, 0, 0, 0]$. FMP-AARS and FMT-AARS reductive property are shown in Table 5.

**Table 5.** FMP-AARS and FMT-AARS reductive property in Class 1

| FMP-AARS premise $A^*(x)$ | FMP-AARS Conclusion and Reductive Property | | |
|---|---|---|---|
| | Conclusion $B^*(y)$ | | Reductive Property |
| [1, 0.3, 0, 0, 0] | more or less form | [0, 0, 0, 0.3, 1] | 100(%) |
| | reduction form | [0, 0, 0, 0.3, 1] | 100(%) |
| [1, 0.09, 0, 0, 0] | more or less form | [0, 0, 0, 0.328, 1] | 95.24(%) |
| | reduction form | [0, 0, 0, 0.274, 0.914] | 94.60(%) |
| [1, 0.548, 0, 0, 0] | more or less form | [0, 0, 0, 0.333, 1] | 95.71(%) |
| | reduction form | [0, 0, 0, 0.27, 0.9] | 92.45(%) |
| [0, 0.7, 1, 1, 1] | more or less form | [0, 0, 0, 0.574, 1] | 17.47(%) |
| | reduction form | [0, 0, 0, 0.157, 0.523] | 18.68(%) |
| *RPCF-FMP* | *FMP−AARS−more or less form* | | 77.10(%) |
| | *FMP − AARS − reduction form* | | 76.43(%) |
| FMT-AARS Premise $B^*(y)$ | FMT-AARS Conclusion and Reductive Property | | |
| | Conclusion $A^*(x)$ | | Reductive Property |
| [1, 1, 1, 0.7, 0] | more or less form | [1 0.574 0 0 0] | 17.47 % |
| | reduction form | [0.523 0.157 0 0 0] | 18.68 % |
| [1, 1, 1, 0.91, 0] | more or less form | [1 0.581 0 0 0] | 13.41 % |
| | reduction form | [0.517 0.155 0 0 0] | 14.57 % |
| [1, 1, 1, 0.452, 0] | more or less form | [1 0.569 0 0 0] | 17.66 % |
| | reduction form | [0.527 0.158 0 0 0] | 23.57 % |
| [0, 0, 0, 0.3, 1] | more or less form | [1 0.3 0 0 0] | 100 % |
| | reduction form | [1 0.3 0 0 0] | 100 % |
| *RPCF−FMT* | *FMT − AARS − more or less form* | | 37.14 % |
| | *FMT − AARS − reduction form* | | 39.2 % |

When compared with Table 2, FMP-QIP reductive property (79.21 % for all) is more than FMP-AARS one (77.10% for *more or less form* and 76.43% for *reduction form*) in Table 8. FMT-AARS reductive property is shown in Table 5. And then FMT-QIP reductive property (all 47.69 %) is more than FMP-AARS one (37.14% for *more or less form* and 39.2% for *reduction form*) in Table 5. Consequently from Table 2-5, we can see that QIP reductive property is more than AARS one, but, not more than TIP, CRI, respectively.

**Proposition 4.5.** *The reductive property of FMP-AARS and FMT-AARS by Łukasiewicz, Gödel, $R_0$ and Gougen are less than FMP-QIP and FMT-QIP with respect to [2], respectively.*

### 4.5. Checking of Proposed DMM

In this section we illustrate the reductive property of our proposed DMM for FMP and FMT. FMP-DM and FMT-DM conclusions and reductive properties are shown in Table 6.

**Table 6.** FMP-DM and FMT-DM conclusion and reductive property in Class 1

| FMP-DM Premise $A^*(x)$ | FMP-DM-*form* Conclusion and Reductive Property | | |
|---|---|---|---|
| | Conclusion $B^*(y)$ | | Reductive Property |
| [1, 0.3, 0, 0, 0] | $P(+1,0,-1)$ *form* | [0, 0, 0, 0.3, 1] | 100 (%) |
| | $P(+1,-1)$ *form* | [0, 0, 0, 0.3, 1] | 100 (%) |
| [1, 0.09, 0, 0, 0] | $P(+1,0,-1)$ *form* | [0.086, 0, 0.086, 0.36, 1] | 91.16 (%) |
| | $P(+1,-1)$ *form* | [0.158, 0, 0.158, 0.41, 1] | 87.26 (%) |
| [1, 0.548, 0, 0, 0] | $P(+1,0,-1)$ *form* | [0, 0.111, 0, 0.3, 1] | 92.83 (%) |
| | $P(+1,-1)$ *form* | [0, 0, 0, 0.3, 1] | 95.05 (%) |
| [0, 0.7, 1, 1, 1] | $P(+1,0,-1)$ *form* | [0, 0.646, 0.646, 0.752, 1] | 68.25 (%) |
| | $P(+1,-1)$ *form* | [0, 1, 1, 0.84, 0.45] | 68.25 (%) |
| RPCF | $FMP-DM-p(+1,0,-1)$ *form* | | 88.06 (%) |
| | $FMP-DM-p(+1,-1)$ *form* | | 87.64 (%) |
| FMT-DM Premise $B^*(y)$ | FMT-DM-*form* Conclusion and Reductive Property | | |
| | Conclusion $A^*(x)$ | | Reductive Property |
| [1, 1, 1, 0.7, 0] | $P(+1,0,-1)$ *form* | [0, 0.7, 1, 1, 1] | 100.0(%) |
| | $P(+1,-1)$ *form* | [0, 0.7, 1, 1, 1] | 100.0(%) |
| [1, 1, 1, 0.91, 0] | $P(+1,0,-1)$ *form* | [0 0.64 0.914 1 0.914] | 91.16 (%) |
| | $P(+1,-1)$ *form* | [0, 0.7, 1, 1, 1] | 95.80 (%) |
| [1, 1, 1, 0.452, 0] | $P(+1,0,-1)$ *form* | [0, 0.7, 1, 0.889, 1] | 92.83 (%) |
| | $P(+1,-1)$ *form* | [0, 0.7 1, 0.778, 1] | 90.61 (%) |
| [0, 0, 0, 0.3, 1] | $P(+1,0,-1)$ *form* | [0, 0.11, 0, 0.3, 1] | 68.25 (%) |
| | $P(+1,-1)$ *form* | [0.56, 0, 0, 0.13, 1] | 85.51 (%) |
| RPCF | $FMT-DM-p(+1,0,-1)$ *form* | | 88.06 (%) |
| | $FMT-DM-p(+1,-1)$ *form* | | 92.98 (%) |

When compared with Table 2, FMP-QIP reductive property (79.21 % for all 4 implications) is less than FMP-DM one (88.06 % for $P(+1,0,-1)$ *form* and 87.64% for $P(+1,-1)$ *form*), respectively, in Table 6. When compared with FMT-QIP reductive property (47.69% for all 4 implications) is less than FMT-DM one (88.06% for $P(+1,0,-1)$ *form* and 92.98% for $P(+1,0,-1)$ *form*) in Table 6. And when compared with Table 3, FMP-DM and FMT-DM reductive property are very more than FMP-CRI and FMT-CRI, respectively.

**Proposition 4.6.** *The reductive properties of FMP-DM and FMT-DM by $P(+1,0,-1)$ form and $P(+1,-1)$ form are more than FMP-QIP, FMT-QIP, FMP-CRI, and FMT-CRI with respect to [2], respectively.*

### 4.6. Comprehensive Comparisons of CRI, TIP, QIP, AARS and Proposed DMM in Class 1

In this section, 5 different fuzzy reasoning methods, concretely 17 methods are compared for FMP and FMT. The reductive property comparison results of the 17 fuzzy reasoning methods in Class 1 are shown in Table 7.

**Table 7.** Comparisons of CRI, TIP, QIP and proposed DMM with respect to the reductive property in Class 1

| No | in Class 1 Fuzzy Reasoning Method | | $RPCF_{FMP-FR}$ | $RPCF_{FMT-FR}$ | $RPCF_{FR}$ |
|---|---|---|---|---|---|
| 1 | proposed DMM | $P(+1,0,-1)$ *form* | 88.06 % | 88.06 % | 88.06 % |
| 2 | | $P(+1,-1)$ *form* | 87.64 % | 92.98 % | 90.31 % |
| 3 | CRI | Gödel; G | 92.45 % | 61.81 % | 77.131 % |
| 4 | | Gougen; Go | 92.45 % | 61.81 % | 77.131 % |
| 5 | (1975) | Łukasiewicz; L | 88.73 % | 61.81 % | 75.273 % |
| 6 | | $R_0$ | 84.23 % | 61.81 % | 73.023 % |
| 7 | | Zadeh; Rz | 78.38 % | 61.81 % | 70.098 % |
| 8 | TIP (1999) | Gödel; G | 92.45 % | 44.69 % | 68.570 % |
| 9 | | Gougen; Go | 92.45 % | 44.69 % | 68.570 % |
| 10 | | Łukasiewicz | 88.73 % | 44.69 % | 66.711 % |
| 11 | | $R_0$ | 84.23 % | 44.69% | 64.461 % |
| 12 | QIP (2015) | Łukasiewicz | 79.21 % | 47.69 % | 63.450 % |
| 13 | | Gödel; G | 79.21 % | 47.69 % | 63.450 % |
| 14 | | $R_0$ | 79.21 % | 47.69 % | 63.450 % |
| 15 | | Gougen; Go | 79.21 % | 47.69 % | 63.450 % |
| 16 | AARS (1990) | *reduction form* | 76.43 % | 39.20 % | 57.818 % |
| 17 | | *more or less form* | 77.10% | 37.14 % | 57.121 % |

17 methods are 5 implication methods in CRI, 4 implication methods in TIP, 4 implication methods in QIP, 2 similarity methods in AARS, and 2 DMM presented in section 3 of this paper. Analysis for the reductive property of 17 fuzzy reasoning methods can be mentioned as follows. The reductive property of CRI and TIP (from 84.23 % to 92.45 % for FMP) are more than DMM (81.78 % and 82.20 % for FMP). But the reductive property of CRI and TIP (61.81 %, 44.69%, for FMT, respectively) are less than DMM (81.78 % and 82.20 % for FMT). Comprehensively proposed DMM (82.246%) is more than CRI (from 70.098 % to 77.131 %) and TIP (from 64.461 % to 68.570 %) for FMP and FMT, respectively. When compared with AARS, our proposed DMM presented in section 3 is not need to calculation of the similarity measures shown in [8], and its reductive property is improved about 24.776(%) for the average than AARS. This result does not largely differ from that in [7], when our proposed DMM is only excepted. Concretely, the best are proposed DM−$P(+1,0,−1)$ *form* and DM−$P(+1,−1)$ *form*. Next best is CRI-Gödel, and the lowest AARS−*reduction form* and AARS−*more or less form*. When compared with the QIP, TIP, and CRI solutions, our proposed DMM is much closer to the consequents that "$x$ *is small*" for FMP and "$y$ *is not large*" for FMT, with respect to the simultaneous consideration of FMP and FMT. In this paper, we have dealt with the reductive property of some fuzzy reasoning methods, i.e., QIP, TIP, CRI, AARS, and our proposed DMM with respect to the criterion for FMP and FMT shown in Table 1. Concretely the fuzzy reasoning methods dealt in section 4 and 5 in this paper are 17.

Through Fig. 1 we can know that the FMT reductive properties of CRI, TIP, QIP, and AARS are less than FMP one, respectively. FMT reductive property by our method is similar to FMP. Through the experiments we have obtained that, proposed DMM is in accordance with human thinking. Otherwise QIP is less than CIR and TIP with respect to the reductive property. Consequently CRI is better than TIP and QIP as well as AARS with respect to the reductive property. The several results differs from the paper [7]. Because in Example 1 in [7], only Case 1 is considered. However in this paper, we all considered different 8 Cases with respect to the reductive property. From Table 7 we can see that the reductive properties of QIP, TIP, CRI and AARS are not better than DMM, respectively, especially AARS method. And the reductive property of our method is 88.454(%). Consequently, in this paper, the reductive properties about QIP, TIP, CRI, and AARS are improved by our proposed DMM.

**Proposition 4.7.** *From the experiment results, the reductive property ranking of the fuzzy reasoning methods in Class 1 are as follows; DMM, CRI, TIP, QIP, and AARS, respectively.*

**Proposition 4.8.** *From the experiment results, the fuzzy reasoning methods, i.e., CRI, TIP, QIP, and AARS, in Class 1 have some information losses, respectively.*

### 4.7. Comparisons of CRI, TIP, QIP, AARS and Proposed Method in Class 2

In this subsection, we compare and analyze about CRI, TIP, QIP, AARS and proposed method for Class 2. computational process is omitted in this paper. According to [10], for Case 4-2 in Table 1, the given premise for example can be defined as follows, respectively, i.e., $A^* = $ *slightly tilted of* $A = [1, 0.2, 0, 0, 0]$, for FMP, $B^* = $ *slightly tilted of* $B = [0, 0, 0, 0.2, 1]$, for FMT. The reductive properties of five fuzzy reasoning methods for Class 2 are shown in Table 11.

**Table 11.** Comparisons of CRI, TIP, QIP and New DM method with respect to reductive property in Class 2

| No | In Class 2 Fuzzy Reasoning Method | | $CF_{FMP}$ (%) | $CF_{FMT}$ (%) | $CF_{Average}$ (%) |
|---|---|---|---|---|---|
| 1 | CRI (1975) | Rz | 81.35 | 74.30 | 77.83 |
| | | Łukasiewicz | 81.35 | 74.30 | 77.83 |
| | | Gödel | 98.45 | 74.30 | 86.38 |
| | | $R_0$ | 81.35 | 74.30 | 77.83 |
| | | Gougen | 98.45 | 74.30 | 86.38 |

| 2 | TIP (1999) | Łukasiewicz | 94.70 | 25.70 | 62.01 |
|---|---|---|---|---|---|
| | | Gödel | 98.45 | 25.70 | 62.08 |
| | | $R_0$ | 90.20 | 25.70 | 57.95 |
| | | Gougen | 98.45 | 25.70 | 62.08 |
| 3 | QIP (2015) | Łukasiewicz | 97.20 | 25.70 | 61.45 |
| | | Gödel | 97.20 | 25.70 | 61.45 |
| | | $R_0$ | 95.85 | 25.70 | 60.78 |
| | | Gougen | 96.20 | 25.70 | 60.95 |
| 4 | AARS (1990) | more or less | 97.17 | 16.01 | 56.59 |
| | | reduction | 96.10 | 18.40 | 57.25 |
| 5 | Proposed DMM | $P(+1,0,-1)$ form | 93.95 | 96.08 | 95.02 |
| | | $P(+1,-1)$ form | 93.97 | 89.13 | 91.55 |

When compared with QIP, TIP, and CRI in Class 2, their difference can be mentioned as follows. From Table 11, we can see that, for four cases in Class 2, the reductive property of QIP is less than CRI and TIP with respect to FMP and FMT. That is, the best is our proposed method, next best CRI, TIP, QIP and the lowest AARS-reduction form and AARS-more or less form in Class 2, this result is similar as in Class 1.

### 4.8. Comprehensive Analysis of Class 1 and Class 2

In this subsection, we have dealt with the reductive property of some fuzzy reasoning methods, i.e., QIP, TIP, CRI, AARS, and our proposed method with respect to for FMP and FMT about Class 1 and Class 2 shown in Table 1 based on combination of [5, 17]. Concretely fuzzy reasoning methods dealt in this paper are 17 one, for Class 1 and Class 2, respectively. The reductive properties of the 5 fuzzy reasoning methods for Class 1 and Class 2 are comprehensively shown in Fig. 1.

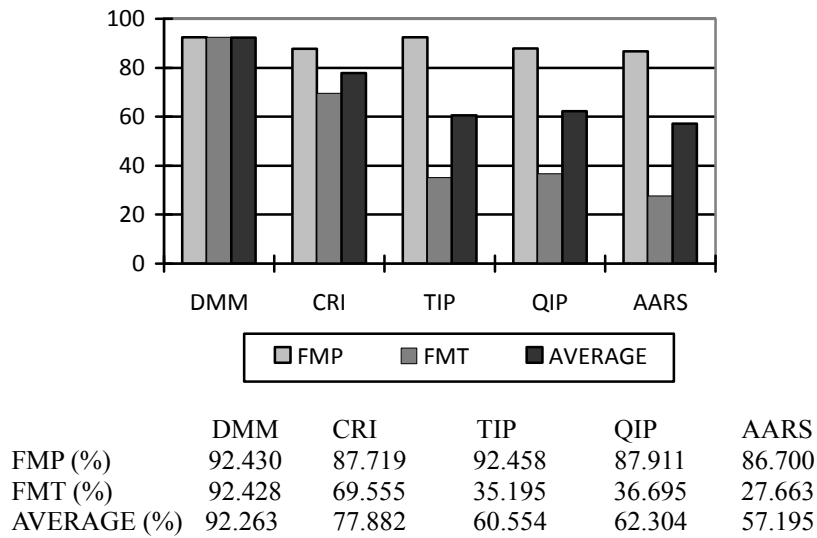

|  | DMM | CRI | TIP | QIP | AARS |
|---|---|---|---|---|---|
| FMP (%) | 92.430 | 87.719 | 92.458 | 87.911 | 86.700 |
| FMT (%) | 92.428 | 69.555 | 35.195 | 36.695 | 27.663 |
| AVERAGE (%) | 92.263 | 77.882 | 60.554 | 62.304 | 57.195 |

Fig. 1. The comprehensive reductive properties of the 5 fuzzy reasoning methods for Class 1 and Class 2

Through the experiments we have obtained that, proposed FMP-DM and FMT-DM methods are in accordance with human thinking. Otherwise QIP is less than TIP and CIR with respect to the reductive property. Consequently CRI (1975) is better than TIP (1999) and QIP (2015) as well as AARS (1990) with respect to the reductive property. These results in Class 1 and Class 2 does not in accordance with the paper [5, 17]. Because Example 1 and 2 in [33] is only considered Case 1 for Class 1 and Case 9 for Class 2. However in this paper we considered all the different four cases for FMP and FMT with respect to the reductive property according to [5, 17]. From Table 10 and 11, we can see that the reductive properties for FMT of QIP, TIP, CRI and AARS are not better than one for FMP, respectively, especially AARS method. And the average reductive property of our method in Class 1 is 89.195(%) and in Class 2 is 93.285(%). In other words the reductive property of our method in Class 2 is better than one in Class 1. Because in Class 2 the given premises of Case 5 and Case 10 are $A^* =$ *slightly tilted of* $A$ for FMP and $B^* =$ *slightly tilted of* $B$ for FMT. Consequently, in this paper, the reductive properties of Examples shown in [33] are extended and improved by our method, for Class 1 and Class 2, respectively.

### 6. Conclusions

In this paper our research results can be summarized as follows.

Firstly, we newly proposed reductive property criterion function for checking of the fuzzy reasoning result. And then, unlike well-known fuzzy reasoning methods based on the similarity measure, we proposed a principle of new fuzzy reasoning method, i.e., FMP-DM and FMT-DM based on distance measure, for short, DMM, and then presented two theorem for FMP and FMT. Through the several experiments, we can know that proposed method is simple and effective.

Secondly, CRI, QIP, TIP and AARS use not only linear operators, i.e., summation, subtraction, product, and division but also nonlinear operators, i.e., max and min, thus they have the information loss in fuzzy reasoning. Especially many nonlinear operators are more used in QIP than TIP and CRI. Therefore QIP is less than TIP and CRI with respect to the reductive property. Otherwise our proposed method uses linear operators, which has not the information loss in fuzzy reasoning. Thereby proposed DMM is more than CRI, QIP, TIP and AARS with respect to the reductive property, respectively.

Thirdly, the Quintuple Implication Principle, i.e., QIP proposed in the paper [33] is correct, but its reductive property, i.e., discovery in their paper is insufficient from viewpoints to [17], so we have checked and recalculated omitting in their paper. From experiment results we have obtained that CRI is better than QIP and TIP as well as AARS for FMP and FMT with respect to the reductive property.

Fourthly, we presented that FMP-AARS and FMT-AARS based on the similarity shown in [8] are not better than FMP-QIP and FMT-QIP in [33], and in accordance with human thinking, respectively. And we have shown that the best is our proposed DM method, next best CRI, and the lowest AARS-reduction and AARS-more or less, and CRI presented in [31] is better than TIP presented in [3] and QIP presented in [33] as well as AARS presented in [8] with respect to the reductive property presented in [17]. This result is the first in this paper, totally different from the result presented in [33].

Fifthly, we compared 17 fuzzy reasoning methods for FMP and FMT. Consequently our proposed DMM is illustratively better than AARS, TIP, and QIP as well as CRI with respect to the reductive property, and in accordance with human thinking.

**SON-IL KWAK**, Corresponding Author, COLLEGE OF INFORMATION SCIENCE, **KIM ILSUNG** UNIVERSITY, PYONGYANG, D P R OF KOREA, *E-mail Address*: *ryongnam18@yahoo.com*

**GUM-JU KIM**, COLLEGE OF INFORMATION SCIENCE, **KIM ILSUNG** UNIVERSITY, PYONGYANG, D P R OF KOREA
*E-mail Address*: **cioc3@ryongnamsan.edu.kp**

**Michio Sugeno,** Corresponding Author, Department of Systems Science, Tokyo Institute of Technology, 4259 Nagatsuta, Midori-ku, Yokohama 227, Japan
*E-mail Address: michio.sugeno@softcomputing.es*

**GWANG-CHOL LI,** COLLEGE OF INFORMATION SCIENCE, **KIM ILSUNG** UNIVERSITY, PYONGYANG, D P R OF KOREA
*E-mail Address*: **cioc4@ryongnamsan.edu.kp**

**MYONG-SUK SON,** COLLEGE OF INFORMATION SCIENCE, **KIM ILSUNG** UNIVERSITY, PYONGYANG, D P R OF KOREA
*E-mail Address*: **cioc5@ryongnamsan.edu.kp**

**HYOK-CHOL KIM,** COLLEGE OF INFORMATION SCIENCE, **KIM ILSUNG** UNIVERSITY, PYONGYANG, D P R OF KOREA
*E-mail Address*: **cioc6@ryongnamsan.edu.kp**

**UN-HA KIM,** COLLEGE OF INFORMATION SCIENCE, **KIM ILSUNG** UNIVERSITY, PYONGYANG, D P R OF KOREA
*E-mail Address*: **cioc7@ryongnamsan.edu.kp**